# INNOVATIVE BERT-BASED RERANKING LANGUAGE MODELS FOR SPEECH RECOGNITION

*Shih-Hsuan Chiu and Berlin Chen*

National Taiwan Normal University, Taipei, Taiwan

{shchiu, berlin}@ntnu.edu.tw

## ABSTRACT

More recently, Bidirectional Encoder Representations from Transformers (BERT) was proposed and has achieved impressive success on many natural language processing (NLP) tasks such as question answering and language understanding, due mainly to its effective pre-training then fine-tuning paradigm as well as strong local contextual modeling ability. In view of the above, this paper presents a novel instantiation of the BERT-based contextualized language models (LMs) for use in reranking of $N$-best hypotheses produced by automatic speech recognition (ASR). To this end, we frame $N$-best hypothesis reranking with BERT as a prediction problem, which aims to predict the oracle hypothesis that has the lowest word error rate (WER) given the $N$-best hypotheses (denoted by PBERT). In particular, we also explore to capitalize on task-specific global topic information in an unsupervised manner to assist PBERT in $N$-best hypothesis reranking (denoted by TPBERT). Extensive experiments conducted on the AMI benchmark corpus demonstrate the effectiveness and feasibility of our methods in comparison to the conventional autoregressive models like the recurrent neural network (RNN) and a recently proposed method that employed BERT to compute pseudo-log-likelihood (PLL) scores for $N$-best hypothesis reranking.

## 1. INTRODUCTION

Due to the dramatic progress in automatic speech recognition (ASR) with various sophisticated deep neural network (DNN) modeling techniques, alongside the availability of large amounts of training data and powerful computational resources, there has been widespread adoption of ASR solutions in many daily life applications such as virtual assistants, smart speakers, interactive voice responses (IVR) and among others [1][2]. However, the performance of ASR in many real-world scenarios is still far from perfect [2][3][4]. To remedy this problem, a lightweight treatment is to rerank (or rescore) the $N$-best hypotheses generated by ASR with more elaborate autoregressive language models, such as high-order $n$-gram language models [5][6][7], the recurrent neural network (RNN) and its improvements with long short-term memory (LSTM) units [8][9][10][11], as well as extra linguistic or acoustic information clues. By doing so, we can avoid any modification of the modules of an ASR system and thus have fast experimental turnover. Notably, the family of RNN-based language models have shown significant and consistent superiority across many ASR hypothesis reranking tasks in relation to the conventional high-order $n$-gram language models, partly attributed to their better ability of capturing longer dependence relationships among words.

In the recent past, a novel neural network-based contextualized language model, viz. Bidirectional Encoder Representations from Transformers (BERT) [12], has been proposed and achieved unprecedented success in a wide array of natural language processing (NLP) tasks such as question answering [13] and language understanding [14]. This can be attributed to its flexible pre-training then fine-tuning paradigm as well as excellent local contextual modeling ability. Nevertheless, as far as we are concerned, BERT has not been sufficiently and systematically studied in the context of ASR $N$-best hypothesis reranking. Building on these observations, this paper presents a novel instantiation of BERT-based contextualized language models for use in reranking of $N$-best hypotheses produced by ASR. To this end, we formulate $N$-best hypothesis reranking with BERT as a prediction problem, which aims to predict the oracle hypothesis that has the lowest word error rate (WER) given the $N$-best hypotheses as the input (denoted by PBERT). Notably, we also explore to leverage task-specific topic information in an unsupervised manner to supplement PBERT in the reranking process (denoted by TPBERT). To the best of our knowledge, TPBERT is the first to integrate unsupervised topic modeling, which captures the global topic information about the task of interest, into a BERT-based method for ASR hypothesis reranking.

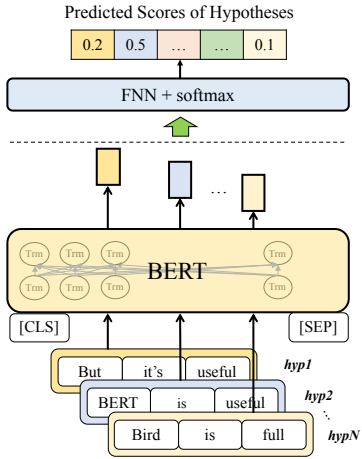

Figure 1: A schematic depiction of PBERT for ASR *N*-best hypothesis reranking.

## 2. RELATED WORK

In this section, we briefly review some representative approaches to *N*-best hypothesis reranking for ASR. Recently, along with the booming of various DNN modeling techniques developed in ASR, the RNN-based LM and its extensions (e.g., LSTM and bidirectional-LSTM [15][16]) have emerged as the de facto standard for ASR *N*-best hypothesis reranking [17][18]. In a follow-up study [19], the authors sought to perform on-the-fly adaptation of the RNN-based LM in an autoregressive manner by augmenting its input at each time stamp with a context-aware vector inferred from the partial word sequence before the current time stamp. In a similar vein, the work in [20] explored a topic modeling approach to extract topic features as additional inputs to the RNN-based LM for genre and topic adaptation on a multi-genre broadcast transcription task. Among other things, there have also been some research efforts devoted so as to modify the model training strategy [21], retrofit the internal model components of an LSTM-based LM [22], and perform ensemble modeling with a mixture of LSTM-based LMs [23]. More recently, an approach analogous to [23] was explored in [24], which replaced LSTM with vanilla Transformer for bidirectionally contextual modeling.

Although the aforementioned LM methods performs well in ASR *N*-best hypothesis reranking, their objectives of model training are all centered around next word prediction (namely, reading a list of words sequentially and predict an upcoming word at each time stamp) but are not closely linked to the reranking accuracy optimization. In this light, a discriminative modeling framework for *N*-best hypothesis reranking was proposed in [25], which consists of an LSTM-based encoder network followed by a binary classification network built on a fully-connected feedforward neural network (FNN).

With the above background, in this paper we put forward a novel *N*-best hypothesis reranking framework, which has at least two noteworthy extensions to the existing state-of-the-art, mainstream methods. First, we leverage BERT as an encoder to simultaneously generate the semantic embeddings of all *N*-best hypotheses, instead of using LSTM to generate the embedding of each hypothesis in a one-by-one or pairwise manner. This way, the embedding of each hypothesis is a function of the entire *N*-best list, with both intra- and inter-hypothesis word-dependence relations (as well as sentence structures) taken into consideration. Second, the embeddings of all *N*-best hypotheses are then fed into a fully-connected FNN at the same time (instead of in a pairwise manner as [25]), while these embeddings can be further augmented with their corresponding task-specific topic representations as extra features to achieve better prediction of the hypothesis out from the *N*-best list that has the lowest WER.

## 3. PROPOSED RERANKING FRAMEWORK

### 3.1. Fundamentals of BERT

BERT [12] is a neural contextualized language model, which makes effective use of bi-directional self-attention (also called the Transformer [26]) to capture both short- and long-span contextual interaction between the tokens in its input sequences, usually in the form of words or word pieces. In contrast to the traditional embedding methods, the advantage of BERT is that it can produce different context-aware representations for the same word at different locations by considering bi-directional dependence relations of words across sentences. The training of BERT consists of two stages: pre-training and fine-tuning. At the pre-training stage, its model parameters can be estimated on huge volumes of unlabeled training data over different tasks such as the masked language model task and the next (relevant) sentence prediction task. At the fine-tuning stage, the pre-trained BERT model, stacked with an additional single- or multi-layer FNN, can be fine-tuned to work well on many NLP-related tasks when only a very limited amount of supervised task-specific training data is made available.

A bit more terminology: for the masked language model task conducted at the pre-training stage, given a token (e.g., word) sequence $\mathbf{x} = [x_1, ..., x_T]$, BERT constructs $\hat{\mathbf{x}}$ by randomly replacing a proper portion of tokens in $\mathbf{x}$ with a special symbol [MASK] for each of them, and designates the masked tokens collectively be $\bar{\mathbf{x}}$. Let $H_\theta$ denotes a Transformer which maps a length-*T* token sequence $\mathbf{x}$ into a sequence of hidden vectors $H_\theta(\mathbf{x}) = [H_\theta(\mathbf{x})_1, H_\theta(\mathbf{x})_2 ..., H_\theta(\mathbf{x})_T]$, then the pre-training objective function of BERT can be expressed by [12][27]

$$\max_\theta \log P_{\text{BERT}}(\bar{\mathbf{x}}|\hat{\mathbf{x}}, \theta) \approx \sum_{t=1}^{T} m_t \log P_{\text{BERT}}(x_t|\hat{\mathbf{x}}, \theta) \quad (1)$$

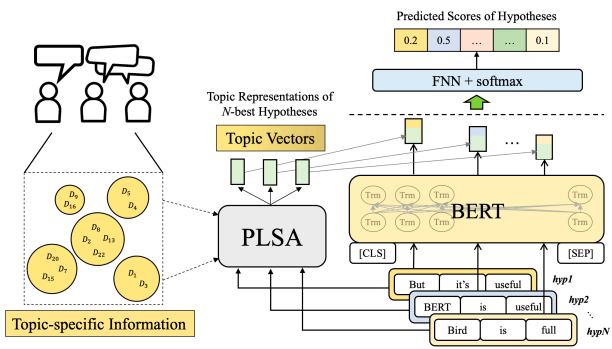

Figure 2: A schematic depiction of TPBERT for ASR *N*-best hypothesis reranking.

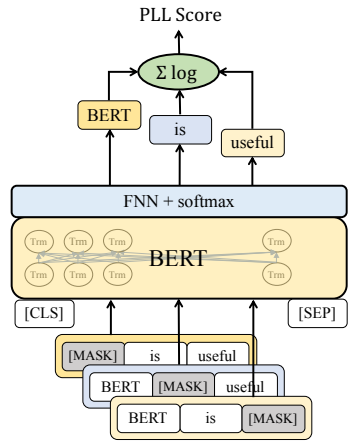

Figure 3: A schematic depiction of MBERT for ASR *N*-best hypothesis reranking.

where $m_t$ is an indicator of whether the token at position $t$ is masked or not.

### 3.2. BERT for ASR *N*-best Hypothesis Reranking

In this paper, we explore a novel use of BERT in the context of ASR *N*-best hypothesis reranking (denoted by PBERT), which the purpose of predicting a hypothesis that would have the lowest WER from an ASR *N*-best hypothesis list. More specifically, PBERT consists of two model components, namely BERT stacked with an additional fully-connected feedforward neural network (FNN), as depicted in Figure 1. For the list of *N*-best hypotheses that is fed into the BERT component, we add [CLS] at the beginning of each hypothesis (regarded as a sentence here) and [SEP] at the end of the hypothesis. In addition, as with the seminal paper of BERT, words involved in each hypothesis will be additionally encoded with a position vector to indicate their positions in the hypothesis, as well as a segment vector to indicate the change of hypotheses if needed. The intermediate goal of PBERT is to use each respective [CLS] vector output from the last layer of the BERT component as the semantic representation of a hypothesis, while all of these [CLS] vectors will be then fed in parallel into the FNN component to output a prediction score (with the softmax normalization) for each hypothesis. Given a set of training utterances, each of which is equipped with an *N*-best hypothesis list generated by ASR and the indication of the oracle hypothesis that has the lowest WER, we can train the FNN component and fin-tune the BERT component as well.

As a side note, the acoustic model score, language model score or their combination score for each hypothesis, obtained from ASR, can be concatenated together with the corresponding [CLS] vector of the hypothesis for feature augmentation.

Table 1: Basic Statistics of the AMI Corpus.

|  | Training Set | Development Set | Evaluation Set |
|---|---|---|---|
| Hours | 78 | 8.71 | 8.97 |
| # Utterances | 108,104 | 13,059 | 12,612 |

### 3.3. Incorporation of Task-specific Topic Information into PBERT

To make the proposed PBERT more task-aware, we seek to add an additional model component to PBERT to empower it to further capture task-specific topic characteristics of *N*-best hypotheses, apart from the both short- and long-span word-level contextual interaction relations that has been already encoded by the BERT component. To this end, we employ probabilistic latent topic analysis (PLSA) [28], a celebrated unsupervised topic modeling method, to extract hypothesis-level topic representation pertaining to the ASR task of interest. In more precise terms, PLSA is first trained in an unsupervised manner with the transcripts of the training utterances so as to globally estimate a set of latent topic distributions that are relevant to the ASR task of interest. At test time, the task-specific topic representation of each hypothesis can be obtained through a simple folding-in process with the set of PLSA-based latent topic distributions estimated beforehand. Thereafter, we refer to this enhanced modeling method of PBERT as TPBERT, whose model structure is schematically depicted in Figure 2. Our TPBERT bears some resemblance to a recently proposed BERT-based model for semantic similarity detection [29].

We remark here that there have been some very recent studies that applied a vanilla BERT-based LM pretrained on a large dataset to score a sentence in terms of the so-called pseudo-log-likelihood (PLL) score, which can be in turn used in conjunction with other scores or embeddings to perform reranking of the *N*-best hypotheses generated by machine translation (MT) or ASR [30][31]. The PLL score of a given sentence (or hypothesis) $\mathbf{x} = [x_1, ..., x_T]$ is defined by

$$\text{PLL}(\mathbf{x}) = \sum_{t=1}^{T} \log P_{\text{BERT}}(x_t | \mathbf{x}_{\setminus x_t}, \theta) \quad (2)$$

where $\mathbf{x}_{\setminus x_t}$ denotes the input sentence **x** that has its word at time stamp $t$ be replaced with the special symbol [MASK]. We denote such a BERT-based scoring method as MBERT, which will be compared to our methods for ASR *N*-best hypothesis reranking in the next section.

## 4. EMPIRICAL EXPERIMENTS

### 4.1. Experimental setup

We evaluate our proposed approach on the AMI meeting transcription database and task [32], while all experiments are conducted using Kaldi toolkit [33]. For the AMI database, the speech corpus consisted of utterances collected with the individual headset microphones (IHM), while a pronunciation lexicon of 50K words was used. Table 1 shows some basic statistics of the AMI corpus for our experiments.

The ASR system employed in this paper was built on the hybrid DNN-HMM paradigm. For acoustic modeling, our recipe was to first estimate a GMM-HMM acoustic model on the training utterances with their corresponding orthographic transcripts, and in turn used the prior distribution of the senones obtained from the GMM components, in conjunction with the LF-MMI training objective, to estimate the time-delay neural network (TDNN) acoustic model [34], which was taken as the seed model. The speech feature vectors were 40 MFCC coefficients extracted from frames of 25 msec length and 10 msec shift, spliced with 100-dimensional i-vectors for speaker adaptation of TDNN. On the other hand, the language model of the baseline ASR system was a trigram language model, which was trained on the transcripts of the AMI training utterances with Kneser-Ney (KN) [5] smoothing.

### 4.2. Experimental Results

In the first set of experiments, we assess the effectiveness of our proposed PBERT and TPBERT methods for ASR *N*-best reranking with two different settings, in comparison to the conventional LSTM and bidirectional-LSTM methods (denoted by LSTM and BLSTM for short, respectively) [15][16]. In Setting (I), we augment the embedding of each hypothesis with the ASR decoding score (a log-liner combination of the acoustic model and language model probability scores). In Setting (II), we first disentangle the decoding score into acoustic model probability score and language model probability score, and then augment the embedding of each hypothesis with these two scores as two additional dimensions. The corresponding results are shown in Table 2, where the WER result of the baseline ASR system is listed for reference. Inspection of Table 2 reveals three noteworthy points. First, PBERT(I) seems to perform at the same level as the LSTM, but does not compete with BLSTM. Second, TPBERT(I) stands out in performance, leading to a relative WER reduction of 2.3% over BLSTM. This also validates the utility of additional incorporation of task-specific topic into PBERT. Third, Setting (II), which disentangles the ASR decoding score into the acoustic model and language model probability scores, appears to worsen the performance when compared Setting (I).

Table 2: The WER results obtained by the proposed PBERT and TPBERT *N*-best reranking methods in comparison to that of the LSTM- and BLSTM-based reranking methods.

| Method | WER (%) |
| --- | --- |
| Baseline ASR | 22.79 |
| LSTM | 21.33 |
| BLSTM | 20.98 |
| PBERT(I) | 21.27 |
| PBERT(II) | 21.69 |
| TPBERT(I) | **20.49** |
| TPBERT(II) | 21.02 |

Table 3: The WER results obtained by variants of the MBERT *N*-best reranking method.

| Method | WER (%) |
| --- | --- |
| MBERT(A) | **20.88** |
| MBERT(B) | 20.98 |
| MBERT(C) | 20.98 |

In the second set of experiments, we evaluate the performance of the recently proposed MBERT method [30][31], so as to further confirm the superiority of our proposed TPBERT method. Note again here that, MBERT treats the vanilla BERT as a masked language model to calculate the PLL score of each hypothesis (*cf.* Section 3.3). Our implementation of MBERT follows the model configuration proposed by [31], with three different training settings: in Setting (A), MBERT simply makes use a publicly-available BERT model that were pretrained on 3.3-billion word tokens [12]; in Setting (B), the BERT model of Setting (A) is further fine-tuned with the ground-truth (reference) transcripts of training utterances; and in Setting (C), the BERT model of Setting (A) is further fine-tuned with the oracle hypotheses of training utterances. Three observations can be drawn from Table 3. First, MBERT(A) delivers a worse WER result than TPBERT(I) and performs slightly better than BLSTM. Second, MBERT(B) and MBERT(C) yield the same WER result, being on par with BLSTM. Third, fine-tuning of the

BERT model bring no benefit to BERT. The above observations do verify the efficacy of TPBERT, which frames the ASR *N*-best rescoring as a prediction problem rather than a language model rescoring problem, such as MBERT that exploits the PLL scores.

## 5. CONCLUSION AND FUTURE WORK

In this paper, we have presented a novel and effective BERT-based method, namely TPBERT, for ASR *N*-best hypothesis reranking, which can additionally leverage task-specific topic information clues to assist the BERT model component in achieving better embeddings of *N*-best hypotheses. A series of empirical evaluations on a benchmark meeting transcription task indeed demonstrate the utility of TPBERT in comparison to some top-of-the-line methods. As to future work, we envisage to explore more sophisticated techniques for task-specific topic or knowledge modeling that can be tightly coupled with our method for use in ASR *N*-best hypothesis reranking, document ranking for spoken document retrieval, and among others [27][35]. We also plan to optimize the model components of our method with some feasible discriminative training criteria [36][37].

## 6. ACKNOWLEDGMENT

This research is supported in part by the Ministry of Science and Technology (MOST), Taiwan, under Grant Number MOST 109-2634-F-008-006- through Pervasive Artificial Intelligence Research (PAIR) Labs, Taiwan, and Grant Numbers MOST 108-2221-E-003-005-MY3 and MOST 109-2221-E-003- 020-MY3. Any findings and implications in the paper do not necessarily reflect those of the sponsors.